\documentclass[runningheads]{llncs}
\usepackage[T1]{fontenc}
\usepackage{graphicx,verbatim}
\usepackage{xcolor}
\usepackage{booktabs} 
\usepackage{multirow}
\usepackage{amsmath,amssymb,amsfonts}
\newcommand{\xct}{{X_{\text{CT}}}}

\newcommand{\xpet}{{X_{\text{PET}}}}

\newcommand{\ect}{{E_{\text{CT}}}}

\newcommand{\tct}{{Z_{\text{CT}}}}
\newcommand{\mct}{{M_{\text{CT}}}}
\newcommand{\epet}{{E_{\text{PET}}}}

\newcommand{\tpet}{{Z_{\text{PET}}}}
\newcommand{\mpet}{{M_{\text{PET}}}}
\newcommand{\dm}{{D^{\text{Cross}}_{{S}}}}
\newcommand{\tm}{{Z_{{S}}}}
\newcommand{\tmcross}{{Z_{{\bar{S}}}}}
\newcommand{\xhatm}{{\hat{X}_{S}}}
\newcommand{\etext}{{E_{\text{TEXT}}}}
\begin{document}

\title{Developing a PET/CT Foundation Model for Cross-Modal Anatomical and Functional Imaging}
%
%\begin{comment}  %% Removed for anonymized MICCAI 2025 submission
\author{
Yujin Oh\inst{1}\textsuperscript{$\dagger$} \and
Robert Seifert\inst{2}\textsuperscript{$\dagger$}\and 
Yihan Cao\inst{2}\and
Christoph Clement\inst{3}\and
Justin Ferdinandus\inst{4}\and
Constantin Lapa\inst{5}\and
Alessandro Liebich\inst{5}\and
Michelle Amon\inst{3}\and
Johanna Enke\inst{5}\and
Sifan Song\inst{1}\and
Runqi Meng\inst{6}\and
Fang Zeng\inst{1}\and
Ning Guo\inst{1}\and
Xiang Li\inst{1}\and
Pedram Heidari\inst{1}\and
Axel Rominger\inst{3}\and
Kuangyu Shi\inst{3}\and
Quanzheng Li\inst{1}\textsuperscript{$\ddagger$}
}

\authorrunning{Y. Oh and R. Seifert et al.}
\institute{
Center for Advanced Medical Computing and Analysis (CAMCA), Harvard Medical School and Massachusetts General Hospital, USA \and
Department of Radiology, Massachusetts General Hospital, USA \and
Department of Nuclear Medicine, Inselspital, University of Bern, Switzerland \and
University Hospital Cologne, University of Cologne, Germany \and
University Hospital Augsburg, Germany\and
Shanghaitech University, China\\
\email{Li.Quanzheng@mgh.harvard.edu}
}

%\end{comment}

%\author{Anonymized Authors}  %% Added for anonymized MICCAI 2025 submission
%\authorrunning{Anonymized Author et al.}
%\institute{Anonymized Affiliations \\
%    \email{email@anonymized.com}}

\maketitle            
\renewcommand{\thefootnote}{\textsuperscript{$\dagger$}}
\footnotetext{These authors contributed equally.}
\renewcommand{\thefootnote}{\textsuperscript{$\ddagger$}}
\footnotetext{Corresponding author.}

\begin{abstract}
In oncology, Positron Emission Tomography-Computed Tomography (PET/CT) is widely used in cancer diagnosis, staging, and treatment monitoring, as it combines anatomical details from CT with functional metabolic activity and molecular marker expression information from PET. However, existing artificial intelligence-driven PET/CT analyses rely predominantly on task-specific models trained from scratch or on limited datasets, limiting their generalizability and robustness. To address this, we propose a foundation model approach specifically designed for multimodal PET/CT imaging. We introduce the Cross-Fraternal Twin Masked Autoencoder (FratMAE), a novel framework that effectively integrates whole-body anatomical and functional or molecular information. FratMAE employs separate Vision Transformer (ViT) encoders for PET and CT scans, along with cross-attention decoders that enable synergistic interactions between modalities during masked autoencoder training. Additionally, it incorporates textual metadata to enhance PET representation learning. By pre-training on PET/CT datasets, FratMAE captures intricate cross-modal relationships and global uptake patterns, achieving superior performance on downstream tasks and demonstrating its potential as a generalizable foundation model.

\end{abstract}

\section{Introduction}

The emergence of foundation models has significantly advanced medical artificial intelligence (AI) in medical image analysis~\cite{ma2024segment,lu2024visual,kim2024transparent,vorontsov2024foundation,bluethgen2024vision}. However, large-scale medical image foundation models have predominantly focused on unimodal imaging modality, such as Computed Tomography (CT) or chest X-ray, leaving a critical gap in their application to multimodal imaging modalities, such like Positron Emission Tomography-Computed Tomography (PET/CT). In specific, PET/CT integrates CT’s detailed anatomical images with PET's functional information or molecular marker expression, making it essential for cancer staging, and treatment monitoring \cite{sadaghiani2021applications,sibille202018f,liu2024systematic}. Despite its clinical importance, there is no dedicated foundation model for PET/CT. \textcolor{black}{The recent success of multimodal AI, which integrates non-imaging modalities such as clinical reports or genetic information, proves the potential of foundation models beyond unimodal imaging \cite{vaidya2025molecular,oh2023llm,zhang2025multimodal,niu2025medical,zhang2024generalist}. Building on this progress, a multimodal PET/CT foundation model can be developed to broaden clinical applications by jointly learning from both PET and CT, such a model can learn richer and more informative representations than traditional unimodal foundation models trained solely on CT. Moreover, leveraging the complementary nature of PET and CT imaging modalites, this approach enhances cross-modal feature learning, leading to more robust and comprehensive medical imaging analysis.} %This requires the need for a multimodal PET/CT foundation model specifically designed to leverage the complementary nature of PET and CT data while ensuring adaptability to diverse clinical applications, even with limited data. %Therefore, current AI-driven approaches on PET/CT analyses rely on limited datasets tailored to specific tracers and tasks, restricting their generalizability and robustness across various radioactive tracers.

\begin{figure}[t!]
\includegraphics[width=\textwidth]{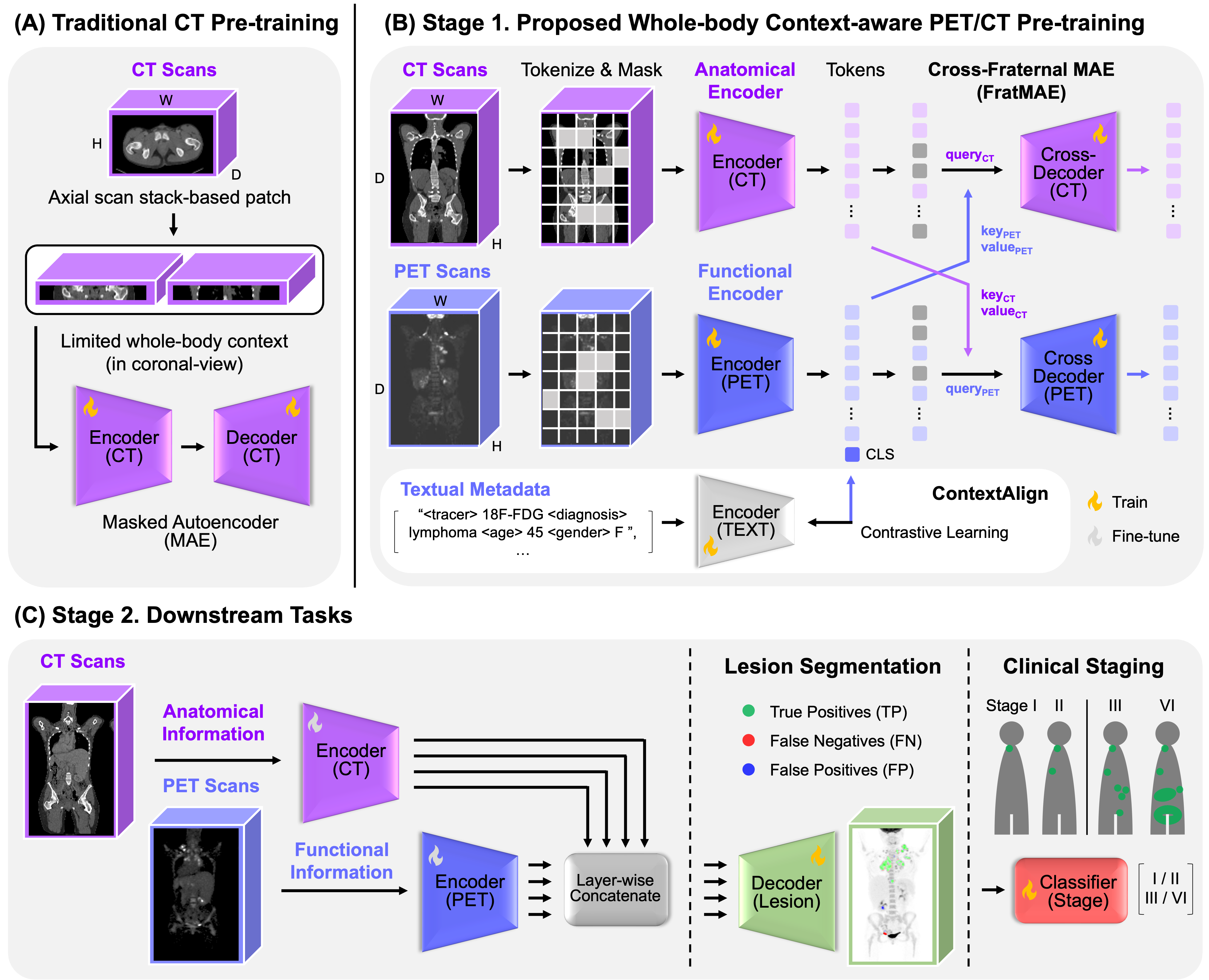}
\caption{(A) Traditional 3D medical image foundation model training. (B) Proposed FratMAE training leveraging coronal scan stacks to capture comprehensive metabolic activity. (C) Downstream tasks include segmentation and clinical staging.} \label{schem}
\end{figure}

When building foundation models for 3D medical imaging, existing training strategies mainly rely on self-supervised learning methods, such as masked autoencoder (MAE)~\cite{he2022masked}. However, these approaches do not explicitly capture the interdependencies between different imaging within multiple modalities, which is essential for PET/CT. Recent advancements including Siamese Masked Autoencoder (SiamMAE)~\cite{gupta2023siamese} and Multi-Modal Masked 3D (M$^3$3D)~\cite{jamal2024m33d}, extend MAE for multimodal learning by leveraging correspondences between different viewpoints. While promising, these methods are designed for different tasks: while SiamMAE primarily focuses on temporal alignment within a single modality, whereas M$^3$3D integrates spatial representations across different dimensional imaging modality. PET/CT, in contrast, involves two inherently distinct imaging modalities—CT, which provides anatomical structure, and PET, which captures functional metabolic activity and molecular marker expression given radioactive tracers. This necessitates a specialized approach that separately encodes each modality while effectively learning their cross-modal relationships.
Furthermore, computational constraints often restrict the training of foundation models for 3D medical imaging to process axial scan stack-based 3D patches \cite{ma2024segment,du2025segvol,koleilat2024medclip,blankemeier2024merlin}, as illustrated in Figure~\ref{schem}(A). While this strategy reduces computational costs and improves spatial understanding of organ structures within axial views, it inherently limits the ability to capture the whole-body context, a crucial factor for learning global uptake patterns of radioactive tracers within PET/CT imaging.

To address these challenges, we propose Cross-Fraternal Twin Masked Autoencoder (FratMAE), a novel PET/CT foundation model that fully integrates whole-body anatomical and functional context. FratMAE employs separate Vision Transformer (ViT) encoders for PET and CT and introduces a cross-attention mechanism in separate ViT decoders, allowing synergistic interactions between modalities during MAE training, as shown in Figure~\ref{schem}(B). Unlike conventional approaches that rely on truncated axial scan stack-based 3D patches, FratMAE processes coronal scan stack-based 3D patches to cover whole-body, allowing the model to learn global uptake patterns of radioactive tracers and their anatomical relevance beyond the limitations of truncated anatomy. 
Additionally, we incorporate textual metadata, such as radiotracer type and basic demographic information, alongside imaging data using a contrastive learning strategy, enhancing PET representation learning. To validate our approach, we pre-train FratMAE on publicly available PET/CT datasets and evaluate its effectiveness on downstream tasks, including lesion segmentation and Hodgkin lymphoma staging, as illustrated in Figure~\ref{schem}(C). Experimental results demonstrate that the FratMAE effectively captures the intricate relationships between anatomical structures, metabolic activity, and textual metadata. Notably, FratMAE achieves robust performance even with limited training data, demonstrating its potential as a scalable and generalizable PET/CT foundation model.

\section{Method}

\subsection{Cross-Fraternal Twin Masked Autoencoder for PET/CT}
Our FratMAE framework builds upon the MAE training mechanism~\cite{he2022masked}, which is designed to reconstruct missing input values from masked tokens, as illustrated in Figure~\ref{schem}(B). To effectively capture cross-modal anatomical and functional correlations, we introduce an asymmetric cross-modal masking strategy that selectively masks one modality while keeping the other unmasked. Furthermore, FratMAE processes whole-body input patches alongside textual metadata that may influence radiotracer uptake patterns in PET imaging.

% \subsubsection{Cross-attention between PET and CT modalites}
Given an input CT image $\xct \in \mathcal{R}^{H \times W \times D}$ and a corresponding PET image $\xpet \in \mathcal{R}^{H \times W \times D}$, where $H,W,D$ denote the height, width, and depth of the input patch, respectively, one modality is randomly selected as the masked target while another is unmasked for each iteration during model training. The masked tokens in the target modality are then reconstructed with the aid of the unmasked modality via a cross-attention mechanism in the decoder. Formally, let $S\in\{\text{CT},\text{PET}\}$ be a randomly chosen modality for masking. The input for each modality encoder can be defined as:
\begin{align}
\tct = \ect(\xct \odot \mct), \quad \tpet = \epet(\xpet \odot \mpet),
\end{align}
where $\ect$ and $\epet$ are the PET and CT encoders, respectively, and the binary masking matrices $\mct, \mpet \in \{0, 1\}^{H \times W \times D}$ follow the asymmetric cross-modal masking strategy. The masking strategy ensures asymmetry that when 50\% of the tokens in the PET modality are masked ($S = \text{PET}$), the CT modality remains fully unmasked ($\mct=\mathbf{1}^{h \times w \times d}$), and vice versa. $\tct, \tpet \in \mathcal{R}^{ch \times h \times w \times d}$ are encoded representations from each encoder, where $h,w,d, ch$  denote the height, width, depth and channel dimension of the encoded representation. Then, each decoder reconstructs the masked tokens:
\begin{align}
\xhatm = \dm(\tm, \tmcross),
\end{align}
where $\dm$ is the cross-attention decoder corresponding to the masked modality $S$. Within each decoder block, the encoded representation of the masked modality $\tm$ serves as the $\bf{query}$, while the encoded representation of the unmasked modality $\tmcross$ serves as both the $\bf{key}$ and $\bf{value}$ in the cross-attention mechanism. This design allows the model to leverage information from the unmasked modality to reconstruct the masked modality’s missing tokens, thereby improving the representation learning of each modality in a complementary manner. Finally, the objective function minimizes the reconstruction loss between the predicted masked tokens and their original values:
\begin{align}\label{mse}
\mathcal{L}_\text{MSE} = \mathbb{E}_{X_S \sim \mathcal{X}_S} \left[ \| X_S - \xhatm \|^2 \right],
\end{align}
where  $X_S$ is sampled from the data distribution $\mathcal{X}_S$, which represents the set of inputs for the chosen modality $S$, and $\| \centerdot \|^2 $ denotes the mean squared error (MSE) loss. By enforcing this asymmetric masking strategy, FratMAE captures cross-modal dependencies while preserving modality-specific information.

\subsection{Whole-body Context Alignment}
To enhance whole-body coverage during foundation model pre-training, PET/CT inputs are first resized to a fixed dimension. Unlike existing foundation models that process patch sizes based on axial scan stacks, we base our scans on coronal slices, which provide better whole-body imaging coverage. Subsequently, 3D patches are randomly cropped with an increased height dimension (by a hight subsampling factor of $k$ relative to the final training patch) and subsampled every $k$ slices along the height axis. For each whole-body patch, we perform contrastive learning to align each PET image representation with its corresponding textual metadata, which includes details such as radiotracer type and basic demographic information. We define this alignment as ContextAlign, and within this module, the textual metadata is encoded using an additional text encoder $\etext$. The alignment is achieved through the InfoNCE loss~\cite{radford2021learning}, defined as: 
\begin{equation}\label{info}
\mathcal{L}_{\text{InfoNCE}} = -\mathbb{E}_{(Z_{\text{PET}}^{\text{CLS}}, Z_{\text{TEXT}}^{\text{CLS}}) \sim p_{\text{data}}} \left[ \log \frac{\exp \left( \frac{Z_{\text{PET}}^{\text{CLS}} \cdot Z_{\text{TEXT}}^{\text{CLS}}}{\tau} \right)}{\sum_{j=1}^{N} \exp \left( \frac{Z_{\text{PET}}^{\text{CLS}} \cdot Z_{\text{TEXT}, j}^{\text{CLS}}}{\tau} \right)} \right],
\end{equation}
where $p_{\text{data}}$ represents the joint distribution of paired PET and text representations, and $Z_{\text{PET}}^{CLS}, Z_{\text{TEXT}}^{CLS} \in \mathcal{R}^{N \times ch}$ denote $\text{CLS}$ tokens from the encoded PET and text representations, respectively. Here, $N$ is the batch size and $ch$ is the representation dimension, and $\tau$ is a temperature hyperparameter.

\subsection{Dataset}

During the pre-training stage, the model is trained on the public AutoPET III dataset~\cite{gatidis2022whole}, comprising 1,292 PET/CT volumes of patients diagnosed with melanoma, lymphoma, lung cancer, prostate cancer, or negative controls. To evaluate the model’s pre-training performance in downstream tasks, we use German Hodgkin Study Group (GHSG) dataset, comprising 515 PET/CT volumes of patients with Hodgkin lymphoma with HD16 (early favorable), HD17 (early unfavorable), and HD18 (advanced stage) trials. Detailed radiotracer and demographic information is provided in Table~\ref{dataset}.

\begin{table}
\caption{Details for dataset characteristics.}\label{dataset}
\resizebox{1\linewidth}{!}{
\begin{tabular}{p{2cm}cc}

\toprule
& \multicolumn{1}{c}{\bf{Stage 1. Pre-training}} & \multicolumn{1}{c}{\bf{Stage 2. Downstream Tasks}} \\
\midrule
{Dataset}  & {AutoPET III} & {German Hodgkin Study Group} \\
\multirow{1}{*}{Radiotracer} & 18F-FDG / 18F-PSMA / 68Ga-PSMA & 18F-FDG \\
\multirow{2}{*}{Diagnosis} &  Melanoma / Lymphoma / Lung Cancer / Prostate Cancer   & \multirow{2}{*}{Hodgkin Lymphoma} \\
& / Negative Control & \\
\multirow{1}{*}{Age} & 61.4 ± 15.6 years& 34.9 ± 12.1 years \\
\multirow{1}{*}{Gender} &  63.7\% male / 36.3\% female& 62.2\% male / 38.8\% female \\

\midrule
Trainset   & 1292 &  \multicolumn{1}{c}{433} \\
\multirow{1}{*}{\shortstack[l]{{Testset }}}   & - & 82 \\

\bottomrule
\end{tabular}
}
\end{table}

\subsection{Implementation details}

\subsubsection{Model Architecture} We adopt the ViT architecture for both CT and PET encoders and decoders, utilizing the standard ViT-B as the backbone  \cite{dosovitskiy2020image} within the MAE framework \cite{he2022masked}. For the text encoder, we employ the CLIP text encoder, also based on the ViT-B architecture \cite{radford2021learning}. The network is trained using 3D patches of 32 $H \times$ 160 $W \times$ 192$D$ voxels and optimized with the AdamW optimizer \cite{loshchilov2017decoupled}, incorporating cosine annealing for learning rate decay.

\subsubsection{Stage 1. Pre-Training} PET/CT inputs are first resized into 160 $H \times$ 160 $W \times$ 192 $D$, with the hight subsampling factor $k$ set to 2. Text metadata includes radiotracer type, diagnosis, age and gender, and they are formatted following the standardized text prompt as, e.g., ``<tracer> 18F FDG <diagnosis> melanoma <age> 61 <sex> M''. The training objective combines MSE loss an InfoNCE loss, as defined in Equation~(\ref{mse}) and (\ref{info}), respectively. The network is trained for 30 epochs with an initial learning rate of $1 \times 10^{-4}$ over 30 epochs, utilizing eight NVIDIA A100 40GB GPUs with a total batch size $N$ of 24. 

\subsubsection{Stage 2. Downstream Tasks} We evaluate our foundation model on two different downstream tasks by retaining the pre-trained encoder weights while integrating a task-specific decoder or classifier module. Prior to inputting the encoder outputs into their respective task-specific components, the encoded representations from the PET and CT encoders are concatenated in a layer-wise manner. To assess the model’s performance under constrained data conditions, we limit the training set to 20\% of the total available dataset.
\begin{itemize}
    \item \textbf{Lesion Segmentation} involves delineating tumor lesions to assist in diagnosis, treatment planning, and prognosis. For training lesion segmentation, we use a convolutional neural network (CNN)-based decoder module, incorporating four intermediate encoder outputs, employs skip connections at multiple levels, following the UNet Transformers (UNETR) architecture \cite{hatamizadeh2022unetr}. To preserve spatial resolution, 3D input patches are randomly cropped from the original scans without resizing. The training objective combines Dice loss and cross-entropy loss for robust optimization. The network is trained for 200 epochs with an initial learning rate of $1 \times 10^{-4}$, utilizing four NVIDIA A100 40GB GPUs with a total batch size of 16.
    \item \textbf{Ann Arbor Staging} classifies Hodgkin lymphoma progression based on the GHSG risk grouping. As a proof of concept, we first group Stage I and II into an early-stage class and Stage III and VI into an advanced-stage class, following GHSG's risk group. The classifier is implemented as a Multi-Layer Perceptron (MLP). To incorporate whole-body context, inputs are resized similarly to the pre-training phase. However, 3D patches are centrally cropped to ensure the inclusion of critical metabolic activity while minimizing peripheral bias. Training is optimized using cross-entropy loss. The network is trained for 500 epochs with an initial learning rate of $5 \times 10^{-5}$, using a single NVIDIA A100 40GB GPU with a total batch size of 2. 
\end{itemize}

\section{Experimental Results}

\subsection{Systemic Comparison on Different Training Approaches}

We first compare the performance of the traditional axial scan stack-based approach with our coronal scan stack-based approach, both without pre-training. As shown in Table~\ref{system}, our approach significantly outperforms the axial scan stack-based method in all tasks. Specifically, it achieves a Dice coefficient of around 0.80 with only 200 training epochs, while the axial scan stack-based approach typically requires over 1,000 epochs for comparable 3D segmentation. For Ann Arbor Staging classification, the axial scan stack-based approach relies on local patch-based feature extraction and necessitates feature aggregation across the body, which limits the encoder’s capacity to model whole-body metabolic interactions and potentially hinders pre-trained encoder's optimal utilization. In contrast, our method directly processes coronal scan stacks, capturing global metabolic interactions and improving staging performance. These results strongly support the adoption of the coronal scan stack as 3D input patches for our foundation model, and demonstrate superiority and adaptability for PET/CT analysis.

\begin{table}[t!]
\caption{Structural comparison on PET/CT imaging analysis.}\label{system}
\resizebox{1\linewidth}{!}{
\begin{tabular}{p{6cm}cccc}

\toprule
Task & \multicolumn{2}{c}{Lesion Segmentation} &\multicolumn{2}{c}{Ann Arbor Staging} \\
Metric& Dice  (CIs$^\dagger$) & IoU (CIs) & Sensitivity (Macro) & Accuracy \\

\midrule
Axial scan stack-based approach & \shortstack[c]{{0.559}{ (0.492-0.623)}} & \shortstack[c]{{0.446}{ (0.384-0.508)}}  & 0.565 & 0.577 \\
Coronal scan stack-based approach &  \shortstack[c]{\bf{0.795}{ (0.770-0.817)}} & \shortstack[c]{\bf{0.672}{ (0.641-0.699)}} & \bf0.615 & \bf0.628\\

\bottomrule
\multicolumn{5}{l}{{{$^\dagger$ CIs denotes 95\% confidence intervals from 1,000-sample non-parametric bootstrapping.}}} \\
\end{tabular}
}
\end{table}

\begin{table}[t!]
\caption{Comparison on downstream tasks with various pre-training settings.}\label{results}
\resizebox{1\linewidth}{!}{
\begin{tabular}{p{2cm}p{4.5cm}cccc}

\toprule
& \multicolumn{1}{c}{\bf{Stage 1. Pre-training}} & \multicolumn{4}{c}{\bf{Stage 2. Downstream Tasks}} \\
\midrule
Task &   &  \multicolumn{2}{c}{Lesion Segmentation} &  \multicolumn{2}{c}{Ann Arbor Staging} \\
\multirow{1}{*}{Metric} & \multirow{1}{*}{}  & Dice (CIs) & IoU (CIs) & Sensitivity (Macro) & Accuracy \\

\midrule
\multirow{5}{*}{\shortstack[l]{{20\% Trainset}}}  & \shortstack[c]{{(1) Baseline}} &  \shortstack[c]{{0.597 (0.553-0.638)}} & \shortstack[c]{{0.454 (0.410-0.493)}} & \shortstack[c]{{0.592 }} & \shortstack[c]{{0.603 }} \\ 
& \shortstack[c]{{(2) MAE}} & \shortstack[c]{{0.534 (0.484-0.578)}} & \shortstack[c]{{0.392 (0.348-0.433)}} & 0.613  & 0.615   \\
& \shortstack[l]{{(3) MAE \quad\enspace\thinspace\! w/ ContextAlign}} & \shortstack[c]{{0.606 (0.561-0.645)}} & \shortstack[c]{{0.462 (0.419-0.502)}} & 0.600 &  0.603  \\
& \shortstack[c]{{(4) FratMAE w/o ContextAlign}} & \shortstack[c]{{0.619 (0.575-0.659)}} & \shortstack[c]{{0.475 (0.434-0.515)}}  & \textcolor{black}{0.570} &    \textcolor{black}{0.577}\\
& \shortstack[l]{{(5) FratMAE }}  & \shortstack[c]{\bf{0.640 (0.599-0.678)}} & \shortstack[c]{\bf{0.496 (0.455-0.534)}} &  \shortstack[c]{\bf{0.648 }} &  \shortstack[c]{\bf{0.654 }} \\

\midrule
\multirow{2}{*}{\shortstack[l]{{40\% Trainset}}}& \shortstack[c]{{(1) Baseline}}  & \shortstack[c]{{0.712 (0.673-0.745)}} & \shortstack[c]{{0.575 (0.534-0.613)}} & \bf0.612 &   \bf0.615\\ 
% & \shortstack[c]{{(2) MAE}} & \shortstack[c]{{0.726 (0.692-0.755)}} & \shortstack[c]{{0.588 (0.551-0.621)}} & 0.563&     0.576\\
% & \shortstack[l]{{(3) MAE \quad\quad+ ContextAlign}}  & & & &     \\
% & \shortstack[c]{{(4) FratMAE}}  & & & &   \\
& \shortstack[l]{{(5) FratMAE}}  & \shortstack[c]{\bf{0.715 (0.676-0.747)}} & \shortstack[c]{\bf{0.578 (0.538-0.614)}} & 0.589 & 0.601\\

\midrule

% \shortstack[l]{{CNN}\\{(Full trainset)}} & \shortstack[c]{{N/A}\\{}\\{}} & & & \shortstack[c]{{0.819 (0.779-0.856)}} & \shortstack[c]{{0.712 (0.666-0.755)}}\\
\multirow{2}{*}{\shortstack[l]{{Full Trainset}}} & (1) Baseline & \shortstack[c]{{0.795 (0.770-0.817)}} & \shortstack[c]{{0.672 (0.641-0.699)}} & 0.615& 0.628\\
% & \shortstack[c]{{(2) MAE}\\{}} & \shortstack[c]{{0.778 (0.748-0.802)}} & \shortstack[c]{{0.652 (0.616-0.682)}}  & & \\
% & \shortstack[l]{{(3) MAE}\\{+ ContextAlign}} & \shortstack[c]{{0.787 (0.763-0.809)}} & \shortstack[c]{{0.661 (0.630-0.689)}} & & \\
% & \shortstack[c]{{(4) FratMAE}\\{}} & \shortstack[c]{{0.772 (0.738-0.797)}} & \shortstack[c]{{0.645 (0.610-0.675)}} & &   \\
& \shortstack[l]{{(5) FratMAE}}  & \shortstack[c]{\bf{0.796 (0.773-0.817)}} & \shortstack[c]{\bf{0.673 (0.643-0.701)}} & \bf0.644 &   \bf0.653\\
% \cmidrule(l){1-1}  \cmidrule(l){2-2}  \cmidrule(lr){3-4}  \cmidrule(lr){5-6} \cmidrule(lr){7-8} \cmidrule(lr){9-10}

\bottomrule
\end{tabular}
}
\end{table}

\subsection{Performance Comparison on Foundation Model Pre-training}

\begin{figure}[t!]
\centering
\includegraphics[width=1\textwidth]{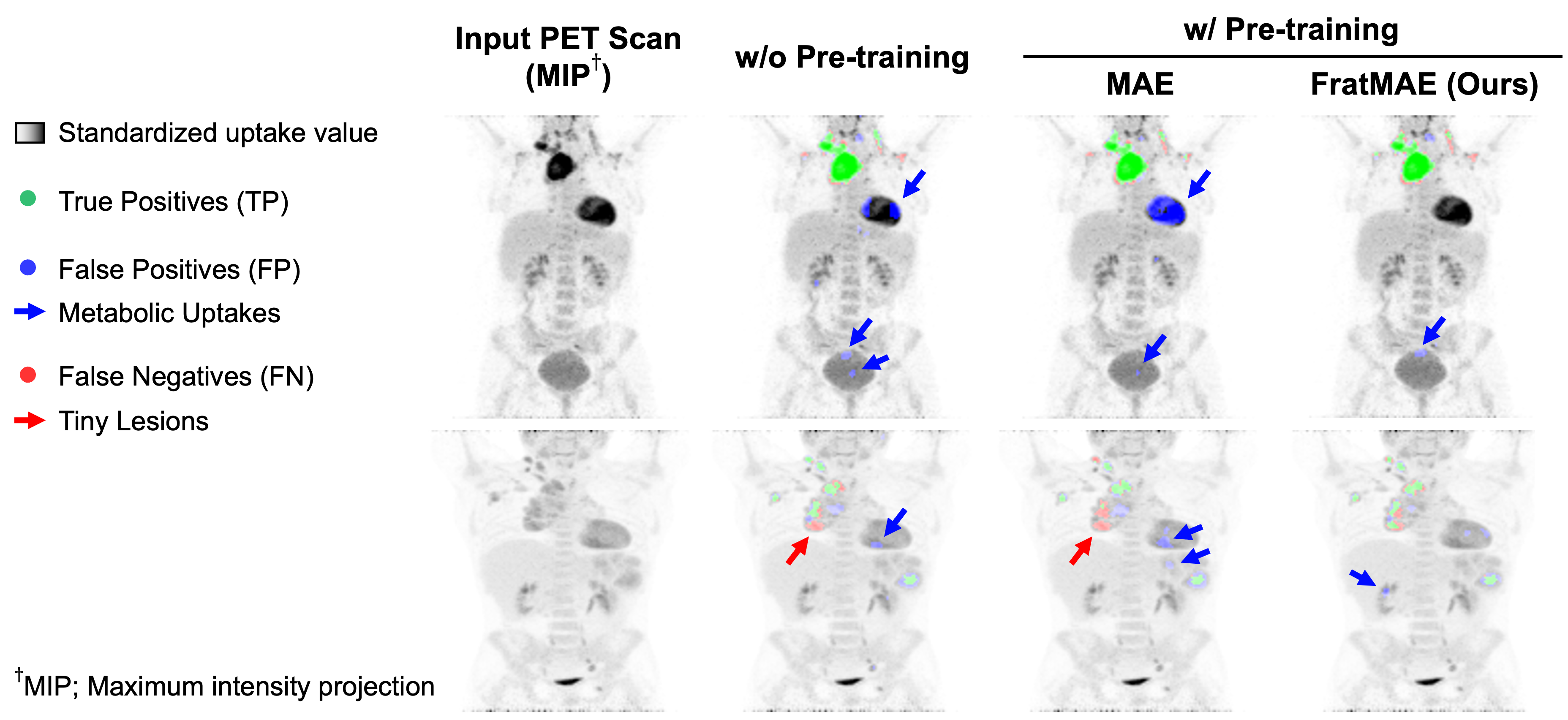}
\caption{ Visual comparison of lesion segmentation using different pre-training methods.} \label{seg}
\end{figure}

We evaluate the efficiency of our foundation model by comparing it on downstream tasks with various pre-training strategies under a limited 20\% training dataset condition. The following approaches are considered: (1) Baseline without pre-training, (2) MAE pre-training, (3) MAE with the ContextAlign module training, (4) FratMAE without ContextAlign module training, and (5) our proposed FratMAE. Table~\ref{results} presents the quantitative results. Our FratMAE consistently outperforms other approaches across both the lesion segmentation and the Ann Arbor staging tasks. Specifically, FratMAE achieves the highest Dice score (0.640) and IoU (0.496) for lesion segmentation, surpassing all baseline methods. Figure~\ref{seg} further provides qualitative comparisons. Notably, models trained without pre-training exhibit a higher incidence of false positives (blue) and false negatives (red). While MAE pre-training mitigates some of these errors, it lacks explicit anatomical-functional alignment during pre-training, limiting its effectiveness in PET/CT analysis. In contrast, FratMAE  leverages cross-modal alignment between anatomical CT information and functional PET information as well as clinical context during pre-training, leading to superior lesion segmentation performance. This results in fewer false positives from metabolic uptakes (blue arrows) and reduced false negatives for tiny lesions (red arrows). Furthermore, it demonstrates superior performance in Ann Arbor staging compared to other methods, proving its effectiveness in learning robust whole-body metabolic representations. We also provide the comparison on (1) Baseline and  (5) FratMAE  under larger trainset conditions (40\% and full), observing a narrowing performance gap. This trend suggests convergence in training; however, given the substantial improvements observed with the 20\% trainset condition, we believe that expanding the pre-training dataset—incorporating multi-center and multi-tracer data—could extend FratMAE's advantages beyond the limited-data regime, further enhancing its generalization capability in PET/CT analysis.

\section{Conclusion}
Our multimodal PET/CT foundation model effectively captures the intricate relationships between anatomical structures, metabolic activity, and text information. It demonstrates strong adaptability in PET/CT analysis tasks, even in data-limited settings. Future work will focus on expanding datasets by incorporating international multi-center and multi-tracer data with richer textual information. Additionally, we aim to extend the scope of downstream tasks to include radiotracer treatment response prediction, further validating the model’s real-world applicability. \textcolor{black}{Lastly, while FratMAE is designed as a multimodal foundation model, it can also be applied independently to CT and PET modalities. We plan to further explore its potential in CT-specific and PET-specific analysis tasks, broadening its impact across medical imaging applications.}

\begin{comment}  
\end{comment}

\clearpage

\bibliographystyle{splncs04}
\bibliography{refs}

\end{document}